\pdfoutput=1
\documentclass[11pt]{article}
\usepackage[final]{acl}
\usepackage{times}
\usepackage{placeins}
\usepackage{latexsym}
\usepackage[T1]{fontenc}
\usepackage[utf8]{inputenc}
\usepackage{microtype}
\usepackage{inconsolata}
\usepackage{longtable}
\usepackage{booktabs}
\usepackage{graphicx}
\usepackage{amsmath}
\usepackage{amssymb}
\usepackage{mathtools}
\usepackage{amsthm}
\usepackage{multirow}
\usepackage{fancyhdr}
\usepackage[capitalize,noabbrev]{cleveref}
\usepackage[textsize=tiny]{todonotes}

\usepackage{fancyhdr}
\fancypagestyle{firstpagefooter}{
  \fancyhf{} 

\fancyfoot[C]{\emph{Proceedings of the 63rd Annual Meeting of the Association for Computational Linguistics} \\ 
July 27, 2025 ©2025 Association for Computational Linguistics}}

\title{Large Language Models are Good Relational Learners}
\author{Fang Wu  \\  Stanford University  \\
  \texttt{fangwu97@stanford.edu} \\\And
  Vijay Prakash Dwivedi\\
  Stanford University  \\
  \texttt{vdwivedi@cs.stanford.edu} \\\And
  Jure Leskovec$^\dagger$  \\
  Stanford University  \\
  \texttt{jure@cs.stanford.edu}}

\begin{document}
\thispagestyle{firstpagefooter}

\maketitle


\begin{abstract}
Large language models (LLMs) have demonstrated remarkable capabilities across various domains, yet their application to relational deep learning (RDL) remains underexplored. Existing approaches adapt LLMs by traversing relational links between entities in a database and converting the structured data into flat text documents. Still, this text-based serialization disregards critical relational structures, introduces redundancy, and often exceeds standard LLM context lengths. We introduce Rel-LLM, a novel architecture that utilizes a graph neural network (GNN)- based encoder to generate structured relational prompts for LLMs within a retrieval-augmented generation (RAG) framework. Unlike traditional text-based serialization approaches, our method preserves the inherent relational structure of databases while enabling LLMs to effectively process and reason over complex entity relationships. 
Specifically, the GNN encoder extracts a local subgraph around an entity to build feature representations that contain relevant entity relationships and temporal dependencies. These representations are transformed into structured prompts using a denormalization process, effectively allowing the LLM to reason over relational structures. Through extensive experiments, we demonstrate that Rel-LLM outperforms existing methods on key RDL tasks, offering a scalable and efficient approach to integrating LLMs with structured data sources. Code is available at~\url{https://github.com/smiles724/Rel-LLM}. 
\end{abstract}

\section{Introduction}
Large language models (LLMs)~\citep{zhao2023survey,minaee2024large}, with their exceptional generalization capabilities in zero or few-shot settings ~\citep{raffel2020exploring, wei2022chain,xu2024retrieval,tang2025medagentsbench,chen2025locagent}, have become foundational tools in diverse areas such as natural language processing~\citep{achiam2023gpt}, computer vision~\citep{liu2024visual}, and information retrieval~\citep{hou2024large}. These advances stem from a series of groundbreaking techniques, including web-scale unsupervised pretraining~\citep{brown2020language}, instruction fine-tuning~\citep{wei2022chain}, and methods to ensure value alignment~\citep{wolf2023fundamental}. 
\begin{figure*}
    \centering
    \includegraphics[width=\linewidth]{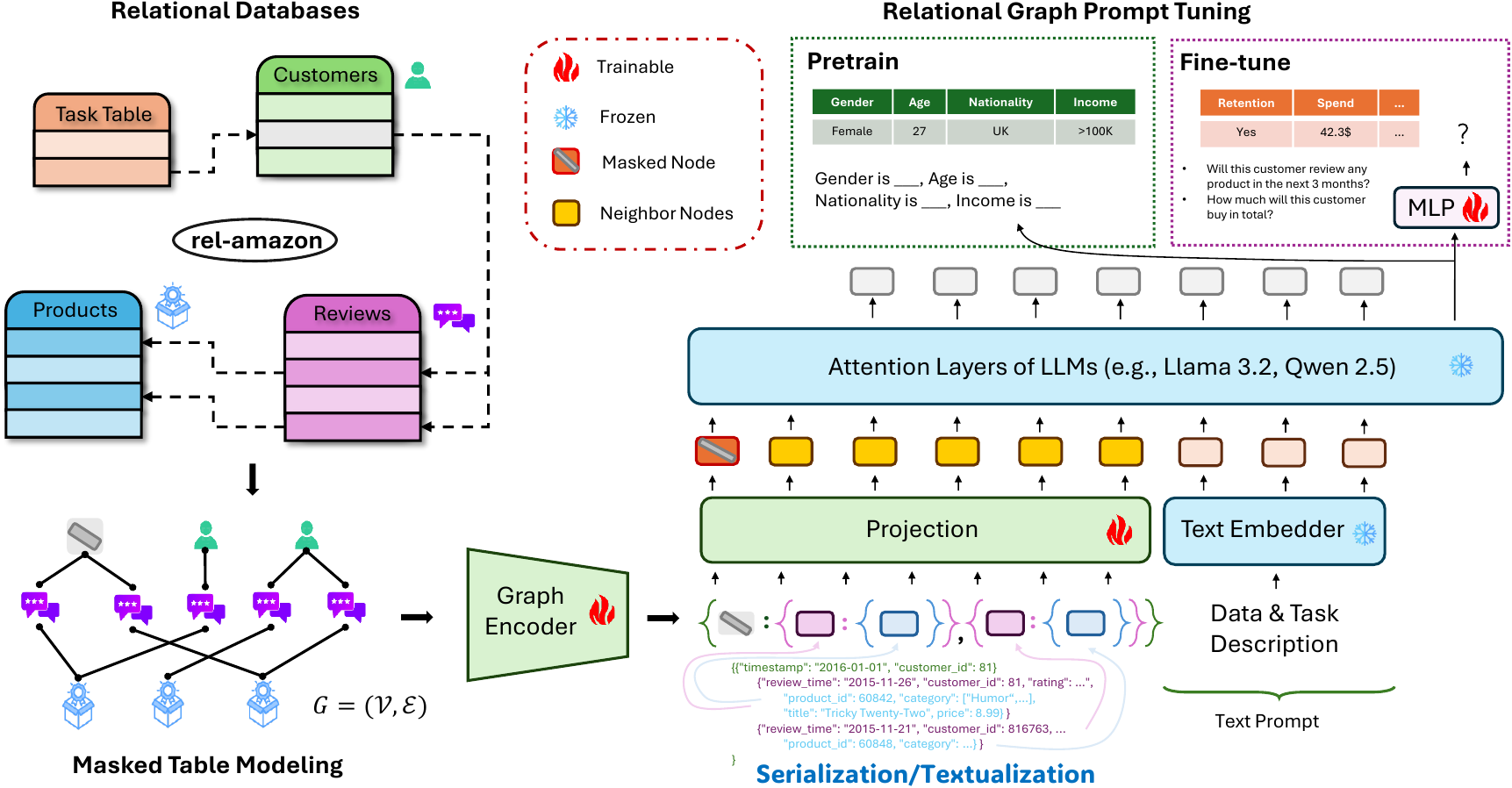}
    \caption{Overview of the Rel-LLM framework. Input from a relational database is fed to an LLM through a graph encoder that captures local connectivity entities in different tables. The graph encoder, along with other trainable components, is pretrained with masked table modeling and fine-tuned for downstream tasks such as customer behavior prediction.}
    \label{fig:model}
\end{figure*}

Despite remarkable achievements, LLMs face challenges in processing and reasoning over complex structured data, such as relational databases~\citep{feyposition,qin2024relational}. Such databases, comprising vast quantities of interconnected data organized into rows and columns, expose LLMs' known restrictions, including hallucinations~\citep{zhang2023siren}, susceptibility to knowledge cutoffs~\citep{gao2023retrieval}, and inefficiencies in capturing explicit relationships in data~\citep {wydmuch2024tackling}. Though retrieval-augmented generation (RAG)~\citep{lewis2020retrieval} has shown promise in alleviating some of these constraints, existing RAG techniques do not model arbitrary connectivity patterns explicitly present in real-world structured data, especially when dealing with relational databases, the backbone of global enterprise data, with approximately 73\% of the world's data~\citep{li2024rllm}.

Relational databases differ fundamentally from single tables due to their interconnected structures. They comprise collections of tables linked by primary and foreign keys, representing one-to-many and many-to-many relationships. This intrinsic complexity poses unique challenges for deep learning (DL), particularly when applying LLMs to leverage the rich semantic and structural information encoded in these links. Recent years have witnessed an increased interest in Relational Table Learning~\citep{dvzeroski2010relational,zahradnik2023deep} to unlock the predictive power of these databases. Furthermore, resources such as GitTables, TabLib~\citep{eggert2023tablib}, and other tabular evaluations~\citep{gardner2024large} have emerged to facilitate this research area. 
Recent benchmarks like CTU~\citep{motl2015ctu}, SJTUTable~\citep{li2024rllm}, and RelBench~\citep{robinson2024relbench} provide datasets for relational tasks, with graph-based approaches~\citep{feyposition,robinson2024relbench} showing promise by representing entities as nodes and relationships as edges.

However, applying LLMs to relational databases remains underexplored. Recent studies~\citep{li2024rllm,wydmuch2024tackling} adapt LLMs by traversing relational links and constructing text-based documents as model inputs. While this aligns with the text-based nature of LLMs, it has significant drawbacks. First, converting relational data into flat text obscures the unique relationships between tables, especially primary-foreign key connections, diminishing LLMs' ability to fully leverage structural insights. Second, the document construction process often results in substantial redundancy, particularly in cases involving nested loop interconnections. This redundancy can lead to repetitive entities appearing multiple times in prompts, complicating the task of extracting key information and increasing computational inefficiency. Finally, real-world relational databases frequently implicate massive datasets with numerous heterogeneous data types and relationships~\citep{fey2023relational}. Processing such large contexts often exceeds the maximum input length of standard LLMs, leading to substantial memory overhead, reduced computational efficiency, and reduced performance~\citep{wang2024beyond}.
These challenges underscore the need for innovative approaches to harness LLMs for relational data processing, driving further research into scalable, structure-aware methodologies capable of unlocking the full potential of relational databases.

In this work, we propose Rel-LLM (see Figure~\ref{fig:model}), a novel framework that combines the structured reasoning capabilities of GNNs with LLMs through retrieval-augmented generation (RAG). 
Rel-LLM uses graph prompt tuning, projecting structured graph embeddings into the LLM’s latent space to guide responses without full fine-tuning. This allows the model to reason over both structured and unstructured data while preserving relational semantics.
Our approach leverages temporal-aware subgraph sampling to ensure causal consistency, a heterogeneous GNN encoder for relational feature extraction \cite{fey2023relational, robinson2024relbench}, and a denormalization-based prompt construction that organizes structured data into a format optimized for LLM processing. 
In addition, Rel-LLM is pretrained using a self-supervised objective that aligns graph-text representations through masked attribute prediction. This pretraining phase ensures that the model can effectively reconstruct entity attributes from corrupted subgraphs, enhancing its ability to reason over relational data and realize zero-shot prediction in low-data regimes.
We evaluate our model on benchmark RDL datasets, demonstrating notable improvements in predictive accuracy, relational reasoning, and temporal consistency compared to existing baselines.

\section{Preliminaries}
\paragraph{Relational Data.}
A relational database~\citep{feyposition} $(\mathcal{T}, \mathcal{L})$ is comprised of a collection of tables $\mathcal{T}=\left\{T_1, \ldots, T_n\right\}$, and links between tables $\mathcal{L} \subseteq$ $\mathcal{T} \times \mathcal{T}$. A link $L=\left(T_{\text{fkey}}, T_{\text{pkey}}\right) \in$ $\mathcal{L}$ between tables exists if a foreign key column in $T_{\text{fkey}}$ points to a primary key column of $T_{\text{pkey}}$. Each table is a set $T=\left\{v_1, \ldots, v_{n_T}\right\}$, whose elements $v_i \in T$ are called rows or entities. Each entity $v \in T$ has four constituent parts $v=\left(p_v, \mathcal{K}_v, x_v, t_v\right)$.

Specifically, $p_v$ is the primary key uniquely identifying the entity $v$. $\mathcal{K}_v \subseteq\left\{p_{v^{\prime}}: v^{\prime} \in T^{\prime} \wedge \left(T, T^{\prime}\right) \in \mathcal{L}\right\}$ are the foreign keys and define links between element $v \in T$ to elements $v^{\prime} \in T^{\prime}$, where $p_{v^{\prime}}$ is the primary key of an entity $v^{\prime}$ in table $T^{\prime}$. $a_v$ is an attribute that holds the entity's information. $t_v$ is an optional timestamp, indicating the time an event occurred. Generally, the attributes in table $T$ contain a tuple of $d_T$ values: $a_v=(a_v^1,...,a_v^{d_T})$. Besides, all entities in the same table have the same columns but values can be absent.

\paragraph{Relational Entity Graph.}
The schema graph (SchG) describes the table-level data structure. Given a relational database $(\mathcal{T}, \mathcal{L})$, $\mathcal{L}^{-1}=\left\{\left(T_{\text{pkey}}, T_{\text{fkey}}\right) \mid\left(T_{\text{fkey}}, T_{\text{pkey}}\right) \in \mathcal{L}\right\}$ denotes its inverse link set. Then, SchG is the graph $(\mathcal{T}, \mathcal{R})$ that arises from the relational database $(\mathcal{T}, \mathcal{L})$, with node set $\mathcal{T}$ and edge set $\mathcal{R}=\mathcal{L} \cup \mathcal{L}^{-1}$. Inverse links ensure that all tables within the SchG are reachable. 

Each relational entity graph (REG) is a heterogeneous graph denoted as $G=(\mathcal{V}, \mathcal{E}, \phi, \psi)$. REG has a node set $\mathcal{V}$, an edge set $\mathcal{E} \subseteq \mathcal{V} \times \mathcal{V}$, node type mapping functions $\phi: \mathcal{V} \rightarrow \mathcal{T}$, and edge type mapping functions $\psi: \mathcal{E} \rightarrow \mathcal{R}$. Here, each node $v \in \mathcal{V}$ belongs to a node type $\phi(v) \in \mathcal{T}$ and each edge $e \in \mathcal{E}$ belongs to an edge type $\psi(e) \in \mathcal{R}$. The sets $\mathcal{T}$ and $\mathcal{R}$ from the SchG define REG's node and edge types.

Given a SchG $(\mathcal{T}, \mathcal{R})$ with tables $T=$ $\left\{v_1, \ldots, v_{n_T}\right\} \in \mathcal{T}$, the node set in REG is defined as the union of all entries in all tables $\mathcal{V}=\bigcup_{T \in \mathcal{T}} T$. Its edge set is then defined as the entity-level pairs that arise from the primary-foreign key relationships in the database, written as:
\begin{equation}
    \mathcal{E}=\left\{\left(v_1, v_2\right) \in \mathcal{V} \times \mathcal{V} \mid p_{v_2} \in \mathcal{K}_{v_1} \text { or } p_{v_1} \in \mathcal{K}_{v_2}\right\}.    
\end{equation}
Moreover, each REG is equipped with three categories of key information. First is type mapping functions $\phi: \mathcal{V} \rightarrow \mathcal{T}$ and $\psi: \mathcal{E} \rightarrow \mathcal{R}$. They map nodes and edges to respective elements of the SchG, making the REG heterogeneous. We set $\phi(v)=T$ for all $v \in T$ and $\psi\left(v_1, v_2\right)=$ $\left(\phi\left(v_1\right), \phi\left(v_2\right)\right) \in \mathcal{R}$ if $\left(v_1, v_2\right) \in \mathcal{E}$.
Second is the time mapping function $\tau: \mathcal{V} \rightarrow \mathcal{D}$. It maps nodes to their timestamp: $\tau: v \mapsto t_v$, which introduces time as a central component and establishes the temporality of the graph. The value $\tau(v)$ denotes the point in time in which the table row $v$ became available or $-\infty$ in the case of non-temporal rows.
The last is the embedding vectors $\mathbf{h}_v \in \mathbb{R}^{d_{\phi(v)}}$ for each $v \in \mathcal{V}$, which contains an embedding vector for each node in the graph. Initial embeddings are obtained via multimodal column encoders~\citep{hu2024pytorch}. 

\section{Methods}
In this section, we introduce Rel-LLM, a novel architecture tailored for RDL, which integrates the strengths of GNNs, LLMs, and RAG. To enable efficient fine-tuning while preserving the LLM’s pretrained language capabilities, we adopt a soft prompting approach by freezing the LLM and conditioning it on the structured outputs of the GNN.

\subsection{Graph Representation Acquisition}
Graph prompt tuning leverages the rich relational structure of databases~\citep{he2024g} to improve downstream predictions by extracting relevant temporal and structural information.

\paragraph{Temporal-aware Subgraph Sampling.} We employ temporal neighbor sampling to construct a subgraph centered around each entity node at a given seed time $t^*$. The seed time $t$ represents the point in history at which a prediction is made. To maintain causality and prevent information leakage, the model exclusively incorporates data from prior to the seed time and ensures no future data is included. During mini-batch training, all nodes within the sampled subgraph are guaranteed to have timestamps earlier than $t^*$~\citep{hamilton2017inductive}. This strategy systematically eliminates temporal leakage and results in a retrieved subgraph denoted as $G^\Diamond = (\mathcal{V}^\Diamond, \mathcal{E}^\Diamond)$.

\paragraph{Graph Encoder.}
To capture the relational structure of $G^\Diamond$, we utilize a heterogeneous variant of the GraphSAGE model~\citep{hamilton2017inductive} with sum-based neighbor aggregation. Given initial node embeddings $\left\{\mathbf{h}_{v_1}^{(0)},...,\mathbf{h}_{v_{n_T}}^{(0)}\right\}$, an $L$-layer GNN iteratively updates the embeddings through message passing, producing $\mathbf{h}_i^{(L)} \in \mathbb{R}^{d_g}$, where $d_g$ denotes the graph encoder’s output dimension. The encoding process is formally defined as:
\begin{equation}
    \mathbf{h}_i^{(L)} = \operatorname{GNN}_{\phi_1}\left(G^\Diamond\right),\quad \mathbf{h}_g^{(L)} =\operatorname{POOL}\left(\mathbf{h}_i^{(L)}\right),
\end{equation}
where POOL represents a mean pooling operation aggregating node embeddings.

\paragraph{Projection Layer.} To align graph embeddings with the LLM’s vector space, we introduce a projection layer using a multilayer perceptron (MLP):
\begin{equation}
    \hat{\mathbf{h}}_i=\operatorname{MLP}_{\phi_2}\left(\mathbf{h}_i^{(L)}\right),\, \hat{\mathbf{h}}_g=\operatorname{MLP}_{\phi_2}\left(\mathbf{h}_g^{(L)}\right) ,
\end{equation}
where $\hat{\mathbf{h}}_i, \hat{\mathbf{h}}_g \in \mathbb{R}^{d_l}$ and $d_l$ corresponds to the hidden dimension of the LLM.

\subsection{Graph Prompt Construction}
Retrieval-augmented systems for LLMs often utilize structured prompts derived from external data sources~\citep{sundar2023ctbls, qin2022external, ye2023large,wu2024insertgnn}. In the context of relational databases, constructing effective prompts is challenging due to heterogeneous tabular data, inconsistent feature scales, and complex interdependencies. Given a relational database $(\mathcal{T}, \mathcal{L})$ and a target prediction task, our approach generates structured documents that encapsulate relevant relational information. Each document is associated with a specific entity $v^*$ from the target table $T^*$ and consists of the following structured components.

\paragraph{Task Context.} Each structured prompt begins with a task description and a question prompt, denoted as $x_{\mathrm{task}}$ and $x_{\mathrm{quest}}$. They detail the database schema, relevant entity relationships, and the objective (e.g., classification or regression). The task description follows templates inspired by prior work~\citep{robinson2024relbench}. 

\paragraph{Denormalization Process}
To preserve the relational structure, we apply a denormalization process to every entity in the graph prompt~\citep{wydmuch2024tackling}. The denormalization follows these recursive steps:
\begin{enumerate}
    \item For the seed entity $v^*$, we recursively follow links from its primary key $\top^*_{\text{pkey}}\in T_{\text{pkey}}$ to foreign keys $T_{\text{fkey}}$, selecting up to $n_{\text{nest}}$ entities from each joined table in a breadth-first manner, up to a recursion depth of $\zeta$.
    \item Aggregate the graph node embedding of all linked entities $\left\{\left\{\hat{\mathbf{h}}_{i,j}\right\}_{i=1}^{n_{\text{nest}}}\right\}_{j=1}^{\zeta}$ into the document representation.
    \item Avoid redundant inclusion by skipping tables already visited in prior denormalization steps.
\end{enumerate}
Efficient execution of this process is ensured by leveraging hash indexes on all primary and foreign keys, allowing rapid retrieval.

\paragraph{Serialization Format}
Designing an effective prompt is a non-trivial task, and many research topics have branched out from prompt engineering alone~\citep{fang2024large}. Here, the fully denormalized entity representation is serialized as a JSON object:
\begin{itemize}
    \item Each entity $v_i$ is represented as a graph neural object $\hat{\mathbf{h}}_i$ reflecting its attributes.
    \item Linked entities appear as nested structures within the parent entity $v_i$ as $\{v_i:\{v_{i,1},...,v_{i,n_{\text{nest}}}\}\}$, where $({\top_{\text{pkey}}}_i, {\top_{\text{fkey}}}_{i,j})\in\mathcal{L}$ for $j=1,...,n_{\text{nest}}$. This reduces the need for multi-hop inference.
\end{itemize}
This leads to the complete graph prompt:
\begin{equation}
    \mathbf{H}^* = \{\hat{\mathbf{h}}^*:\{\{\hat{\mathbf{h}}_{i}^*: \{\hat{\mathbf{h}}_{i,j}^*: \{...\}\}_{j=1}^{n_{\text{nest}}}\}_{i=1}^{n_{\text{nest}}}\}\},
\end{equation}
where $\mathbf{H}^*$ contains at most $n_{\text{nest}}\zeta$ elements.
The choice of JSON format is motivated by prior findings~\citep{singha2023tabular}, demonstrating its effectiveness in encoding tabular and relational data for LLMs. By structuring input documents in this manner, we enable LLMs to effectively reason over relational data while preserving contextual and temporal integrity. Figure~\ref{fig:context}  in the Appendix provides a clear visualization of this process.

\subsection{Answer Generation}
\paragraph{Text Embedder.}  To leverage the text-reasoning capabilities of LLMs, we transform the task context $x_{\mathrm{task}}$ and the question prompt $x_{\mathrm{quest}}$ to an embedding $\mathbf{h}_{\text{text}}$ using a text embedder, which is the first layer of a pretrained and frozen LLM:
\begin{equation}
\mathbf{h}_{\text{text}}=\operatorname{TextEmbedder}\left(\left[x_{\mathrm{task}} ; x_{\mathrm{quest}}\right]\right),
\end{equation}
where $\mathbf{h}_{\text{text}}\in \mathbb{R}^{n_{\text{text}} \times d_l}$. $[\cdot;\cdot]$ represents the concatenation operation, and $n_{\text{text}}$ is the number of total textual tokens.

\paragraph{LLM Generation with Graph Prompt Tuning.} The final stage involves generating the answer $Y$ given the graph neural prompt $\mathbf{H}^*$, acting as a soft prompt, and the text embedder output $\mathbf{h}_{\text{text}}$. These inputs are fed through the self-attention layers of a pretrained frozen LLM, with parameter $\theta$. The generation process is represented as follows:
\begin{equation}
\begin{split}
    p_{\theta, \phi_1, \phi_2}&\left(Y \mid G^\Diamond, x_q\right)= \\
    &\prod_{i=1}^r p_{\theta, \phi_1, \phi_2}\left(y_i \mid y_{<i},\left[{\mathbf{H}}^* ; \mathbf{h}_{\text{text}}\right]\right),
\end{split}
\end{equation}
where $\left[\mathbf{H}^* ; \mathbf{h}_{\text{text}}\right]$ concatenates the graph prompt token $\mathbf{H}^*$ and the text embedder output $\mathbf{h}_{\text{text}}$. While $\theta$ is frozen, the graph token $\mathbf{H}^*$ receives gradients, enabling the optimization of the parameters of the graph encoder $\phi_1(\cdot)$ and the projection layer $\phi_2(\cdot)$ through standard backpropagation.

\paragraph{Answer Generation Strategy.} We explore three sorts of answer generation methods, including plain text, token distribution, and MLP transformation. 
\begin{itemize}
    \item  Plain text generation directly outputs a sequence of tokens forming the human-readable text. It is direct and interpretable but lacks richness for complex tasks. 
    \item Token distribution outputs a distribution over possible tokens, which can be useful for probabilistic or multi-modal tasks.
    \item MLP transformation applies a lightweight neural network to refine or project the LLMs’ latent representations into task-specific spaces.
\end{itemize}
Empirical evaluations on RelBench reveal that different tasks benefit from distinct strategies, underscoring the importance of selecting an appropriate answer generation method based on task-specific requirements.

\subsection{Pretraining}
We design a pretraining objective to align the graph-text representations through self-supervised attribute prediction. 

\paragraph{Masked Subgraph Construction.} We randomly select a subset of nodes $\mathcal{V}_{\text {mask }} \subseteq \mathcal{V}^{\diamond}$ with probability $p_{\text {mask }}$. Then for each masked entity $v_{\mathrm{mask}}\in \mathcal{V}_{\text {mask}}$, we build a temporal-aware subgraph $G^\Diamond_{\text{mask}}$ as described before. Besides, the raw feature of those masked entities $v_{\mathrm{mask}}$ is replaced with a learnable mask token $\mathbf{h}_{\text{mask}} \in$ $\mathbb{R}^{d_g}$, creating a corrupted subgraph.

\paragraph{Graph Encoding with Noise.} We forward and process $G_{\text{mask}}^{\diamond}$ through the graph encoder and projection layer:
\begin{equation}
    \hat{\mathbf{h}}_{\text{mask}} = \operatorname{MLP}{\phi_2}\left(\operatorname{GNN}{\phi_1}\left(G^\Diamond_{\text{mask}}\right)\right).
\end{equation}

\paragraph{Masked Attribute Prediction.} After obtaining $\hat{\mathbf{h}}_{\text{mask}}$ for each masked node $v_i \in \mathcal{V}_{\text{mask}}$, we require the LLM to reconstruct its original attributes through the plain text generation. Let $A_i=\left\{\left(k_j, a_j\right)\right\}_{j=1}^{d^T}$ denote the column-value pairs of $v_i$ 's attributes. We:
\begin{enumerate}
\item Apply random permutation $\pi$ to the column order: $A_i^\pi=\left\{\left(k_{\pi(j)}, a_{\pi(j)}\right)\right\}_{j=1}^m$
\item Format the target sequence as $Y_i=$"$k_{\pi(1)}$ is $a_{\pi(1)}, \ldots , k_{\pi(m)}$ is $a_{\pi(m)}$" and the masked prompt template as $X_i=$"$k_{\pi(1)}$ is $\mathrm{[MASK]}, \ldots , k_{\pi(m)}$ is $\mathrm{[MASK]}$", where $\mathrm{[MASK]}$ is the mask token of LLMs.  
\end{enumerate}
The pretraining loss is computed as:
\begin{equation}
\begin{split}
    \mathcal{L}_{\text{pretrain}}&(\phi_1,\phi_2, \mathbf{h}_{\text{mask}}) = -\frac{1}{|\mathcal{V}_{\text{mask}}|}\\
    & \sum_{v_i \in \mathcal{V}_{\text{mask}}} \sum_{t=1}^{|Y_i|} \log p_\theta\left(y_i^{(t)} \mid y_i^{(<t)}, \hat{\mathbf{h}}_{\text{mask}}\right)
\end{split}
\end{equation}
where $y_i^{(t)}$ is the $t$-th token in $Y_i$. This objective trains $\phi_1, \phi_2$, and $\mathbf{h}_{\text{mask}}$ to encode graph structures into text-compatible representations that preserve attribute semantics.
\begin{table*}[t]
    \caption{Entity classification results (AUROC, higher is better) on RELBENCH. The best and the second best values are in bold and underlined, respectively. \emph{Rel-Zero} records the zero-shot performance of our methods, while \emph{Rel-LLM} corresponds to the performance after fine-tuning. Standard deviations are reported in Appendix Table~\ref{tab:std_cls}. }   
    \centering
    \resizebox{1.75\columnwidth}{!}{
    \begin{tabular}{ccccccc|cc} \toprule
     \textbf{Dataset} & \textbf{Task} & \textbf{Split} & \textbf{LightGBM} & \textbf{RDL}  & \textbf{ICL} & \textbf{ICL + MLP} & \textbf{Rel-Zero} & \textbf{Rel-LLM}  \\ \midrule
     \multirow{4}{*}{rel-amazon} & \multirow{2}{*}{user-churn} & Val & 52.05 & \underline{70.45} & -- -- & -- -- & 59.97 & \textbf{71.85} \\
     & & Test & 52.22 & \underline{70.42} & 60.56 & 66.56 & 60.07 &  \textbf{71.89} \\
     & \multirow{2}{*}{item-churn} & Val & 62.39 & \underline{82.39} & -- -- & -- -- & 64.21 & \textbf{82.97} \\
     & & Test & 62.54 & \underline{82.81} & 71.96 & 80.16 & 64.10 & \textbf{83.37}  \\  \midrule
     \multirow{4}{*}{rel-avito} & \multirow{2}{*}{user-visits} & Val & 53.31 & \underline{69.65} & -- -- & -- -- &  58.32 & \textbf{70.26} \\
     & & Test & 53.05 & \underline{66.20} & 60.28 & 64.98 & 56.17 &  \textbf{67.01} \\
     & \multirow{2}{*}{user-clicks} & Val & 55.63 & \underline{64.73} & -- -- & -- -- & 58.89 & \textbf{65.25} \\
     & & Test & 53.60 & \underline{65.90} & 61.32 & 71.31 & 62.28 & \textbf{66.74} \\  \midrule 
     \multirow{4}{*}{rel-event} & \multirow{2}{*}{user-repeat} & Val & 67.76 & \underline{71.25} & -- -- & -- -- &  61.93 & \textbf{73.03} \\ 
     & & Test & 68.04 & \underline{76.89} & 76.38 & 76.72 & 68.12 & \textbf{79.26}  \\
     & \multirow{2}{*}{user-ignore} & Val & 87.96 & \textbf{91.70} & -- -- & -- -- & 57.47 & \underline{89.95} \\
     & & Test & 79.93 & {81.62} & 78.55 & \textbf{84.02} & 61.32 & \underline{83.74} \\   \midrule
     \multirow{4}{*}{rel-f1} & \multirow{2}{*}{driver-dnf} & Val & 68.42 & \underline{71.36} & -- -- & -- -- & 70.35 & \textbf{76.04}  \\
     & & Test & 68.56 & {72.62} & 65.81 & \textbf{78.41} & 71.84 & \underline{77.15} \\
     & \multirow{2}{*}{driver-top3} & Val & 67.76 & \underline{77.64} & -- -- & -- -- & 58.57 & \textbf{78.03}\\
     & & Test & 73.92 & 75.54 & \textbf{88.47} & \underline{87.36} & 70.64 &  82.22\\   \midrule
     \multirow{2}{*}{rel-hm} & \multirow{2}{*}{user-churn} & Val & 56.05 & \underline{70.42} & -- -- & -- -- & 56.83 & \textbf{70.89 }\\
     & & Test & 55.21 & \underline{69.88} & 64.34 &  68.72 & 55.95 & \textbf{70.55} \\  \midrule
     \multirow{4}{*}{rel-stack} & \multirow{2}{*}{user-engagement} & Val & 65.12 & \underline{90.21} & -- -- & -- -- &  69.97 & \textbf{90.75} \\
     & & Test & 63.39 & \underline{90.59} & 81.01 & 87.09 & 69.46 & \textbf{91.21} \\
     & \multirow{2}{*}{user-badge} & Val & 65.39 & \underline{89.86} & -- -- & -- -- & 64.47 & \textbf{90.80} \\
     & & Test & 63.43 & \underline{88.86} & 71.13 & 88.19 & 62.12 &  \textbf{89.64} \\  \midrule
     \multirow{2}{*}{rel-trial} & \multirow{2}{*}{study-outcome} & Val & 68.30 & \underline{68.18} & -- -- & -- -- &  58.23 & \textbf{70.93} \\
     & & Test & 70.09 & \underline{68.60} & 55.72 & 68.38 & 59.02 & \textbf{71.04}  \\  \midrule
     \multicolumn{2}{c}{\multirow{2}{*}{\textbf{Average}}} & Val & 64.18 & \underline{76.49} & -- -- & -- -- & 61.31 & \textbf{77.56} \\
     & & Test & 63.66 & {75.83} & 69.63 & \underline{76.83} & 63.42 & \textbf{77.82} \\ \bottomrule 
\end{tabular}}
    \label{tab: cls}
\end{table*}
\section{Experiments} 
\subsection{Experimental Setups}
\paragraph{Basic.} We leverage the Llama 3.2-1B~\citep{touvron2023Llama} as $p_\theta$, which supports context-size up to 128k tokens. More details are elaborated in Appendix~\ref{app:exp}.

\paragraph{Dataset and Baselines.} RELBENCH~\citep{robinson2024relbench} contains 7 datasets with 30 predictive tasks each with a rich relational structure, providing a challenging environment for developing and comparing RDL methods. Several baselines are selected, including (1) \textbf{LightGBM} learns a LightGBM~\citep{ke2017lightgbm} regressor over the raw entity features to predict the numerical targets, where only information from the single entity table is used. (2) \textbf{RDL} combines GNN predictive models with deep tabular models that extract initial entity-level representations from raw tables. (3) \textbf{ICL}~\citep{wydmuch2024tackling} leverages the power of LLMs' in-context learning on tabular data. Its variant \textbf{ICL + MLP} tests the strength of LLMs' representations. (4) \textbf{Entity mean/median} calculates the mean/median label value for each entity in training data and predicts the mean/median value for the entity. (5) \textbf{Global mean/median} calculates the global mean/median label value over the training data and predicts the same mean/median value across all entities. (6) \textbf{Global zero} predicts zero for all entities.
\begin{table*}[t]
    \caption{Entity regression results (MAE $\downarrow$) on RELBENCH. The best and suboptimal values are in bold and underlined, separately. Standard deviations are reported in Appendix Table~\ref{tab:std_reg}. }   
    \centering
    \resizebox{1.0\linewidth}{!}{%
    \begin{tabular}{ccccccccccc|c} \toprule
     \textbf{Dataset} & \textbf{Task} & \textbf{Split} & \begin{tabular}{l} \textbf{Global} \\ \textbf{Zero}  \end{tabular} & \begin{tabular}{l}  \textbf{Global} \\ \textbf{Mean} \end{tabular} & \begin{tabular}{l}  \textbf{Global} \\ \textbf{Median}  \end{tabular} & \begin{tabular}{l} \textbf{Entity} \\  \textbf{Mean}  \end{tabular} & \begin{tabular}{l}\textbf{Entity} \\ \textbf{Median} \end{tabular} & \textbf{LightGBM} & \textbf{RDL} & \begin{tabular}{c}\textbf{ICL} \\ \textbf{+ MLP} \end{tabular} & \textbf{Rel-LLM} \\ \midrule
     \multirow{4}{*}{rel-amazon} & \multirow{2}{*}{user-ltv} & Val & 14.141 & 20.740 & 14.141 & 17.685 & 15.978 & 14.141 & \underline{12.132} &  -- -- & \textbf{11.773} \\
     & & Test & 16.783 & 22.121 & 16.783 & 19.055 & 17.423 & 16.783 & \underline{14.313} & 14.864 & \textbf{14.087} \\
     & \multirow{2}{*}{item-ltv} & Val & 72.096 & 78.110 & 59.471 & 80.466 & 68.922 & 55.741 & \underline{45.140} &  -- -- & \textbf{44.167} \\
     & & Test & 77.126 & 81.852 & 64.234 & 78.423 & 66.436 & 60.569 & \underline{50.053} & 52.682 & \textbf{48.224} \\ \midrule
     \multirow{2}{*}{rel-avito} & \multirow{2}{*}{ad-ctr} & Val & 0.048 & 0.048 & 0.040 & 0.044 & 0.044 & 0.037 & \underline{0.037} &  -- -- &  \textbf{0.033} \\
     & & Test & 0.052 & 0.051 & 0.043 & 0.046 & 0.046 & 0.041 & {0.041} & \textbf{0.0036} & \underline{0.037} \\ \midrule
     \multirow{2}{*}{rel-event} & \multirow{2}{*}{user-attendance} & Val & 0.262 & 0.457 & 0.262 & 0.296 & 0.268 & 0.262 & \underline{0.255} &  -- -- & \textbf{0.239} \\
     & & Test & 0.264 & 0.470 & 0.264 & 0.304 & 0.269 & 0.264 & \underline{0.258} & 0.293 & \textbf{0.251}  \\ \midrule 
     \multirow{2}{*}{rel-f1} & \multirow{2}{*}{driver-position} & Val & 11.083 & 4.334 & 4.136 & 7.181 & 7.114 & 3.450 & \underline{3.193} &  -- -- & \textbf{3.050} \\
     & & Test & 11.926 & 4.513 & 4.399 & 8.501 & 8.519 & 4.170 & {4.022} & \textbf{3.539} & \underline{3.967} \\ \midrule
     \multirow{2}{*}{rel-hm} & \multirow{2}{*}{item-sales} & Val & 0.086 & 0.142 & 0.086 & 0.117 & 0.086 & 0.086 & \underline{0.065} &  -- -- & \textbf{0.060} \\
     & & Test & 0.076 & 0.134 & 0.076 & 0.111 & 0.078 & 0.076 & \underline{0.056} & 0.057 & \textbf{0.052} \\ \midrule
     \multirow{2}{*}{rel-stack} & \multirow{2}{*}{post-votes} & Val & 0.062 & 0.146 & 0.062 & 0.102 & 0.064 & 0.062 & \underline{0.059} &  -- -- & \textbf{0.057} \\
     & & Test & 0.068 & 0.149 & 0.068 & 0.106 & 0.069 & 0.068 & \underline{0.065} & 0.090 & \textbf{0.062} \\ \midrule
     \multirow{4}{*}{rel-trial} & \multirow{2}{*}{study-adverse} & Val & 57.083 & 75.008 & 56.786 & 57.083 & 57.083 &  \underline{45.774} & {46.290} &  -- -- & \textbf{45.395}  \\
     & & Test & 57.930 & 73.781 & 57.533 & 57.930 & 57.930 & \underline{44.011} & {44.473} & 51.845 & \textbf{43.682} \\
     & \multirow{2}{*}{site-success} & Val & 0.475 & 0.462 & 0.475 & 0.447 & 0.450 & 0.417 & \underline{0.401} &  -- -- & \textbf{0.397} \\
     & & Test & 0.462 & 0.468 & 0.462 & 0.448 & 0.441 & 0.425 & \underline{0.400} & 0.441 & \textbf{0.397} \\ \midrule
     \multicolumn{2}{c}{\multirow{2}{*}{\textbf{Average}}} & Val & 17.260 & 19.939 & 15.051 & 18.158 & 16.668 & 13.330 & \underline{11.952} &  -- -- & \textbf{11.685} \\
     & & Test & 18.299 & 20.393 & 15.985 & 18.325 & 16.801 & 14.045 & \underline{12.631} & 13.761 & \textbf{12.306} \\ \bottomrule 
    \end{tabular}}
    \label{tab: reg}
\end{table*}
\subsection{Entity Classification}
Table~\ref{tab: cls} presents the results of entity classification on the RELBENCH benchmark, where AUROC is used as the evaluation metric. Our method, Rel-LLM, consistently outperforms or matches all baselines across different datasets and tasks. Notably, Rel-LLM achieves the highest average AUROC of 77.82 on the test set, surpassing RDL (75.83) and outperforming ICL + MLP (76.83) -- a method that relies on an exhaustive search to determine the best combination of document generation parameters.
Rel-LLM particularly excels in datasets where entity behavior follows structured patterns, such as \textsc{rel-f1}, where it achieves state-of-the-art performance. This can be attributed to the fact that Formula 1 drivers and their historical performance are well-documented in large-scale text corpora used for LLM pretraining, allowing the model to leverage pre-existing factual knowledge. However, in other datasets, such as \textsc{rel-amazon} and \textsc{rel-event}, where LLMs cannot directly retrieve pre-trained knowledge, the model must rely on learning from historical interactions -- such as purchase behaviors or user engagement patterns.
A key limitation of ICL-based methods (ICL and ICL + MLP) is their strong dependency on the quality, order, and quantity of in-context examples. Their effectiveness can be significantly constrained by the context length limits of modern LLMs (\emph{e.g.}, 128K tokens for Llama 3.2-1B), making them sensitive to prompt design. In contrast, Rel-LLM eliminates the need for explicitly providing textual in-context examples by incorporating graph-based node embeddings into its reasoning process. This allows Rel-LLM to capture historical behaviors more efficiently and robustly, making it well-suited for modeling complex temporal and relational dependencies in entity classification tasks.

\subsection{Entity Regression}
Entity-level regression tasks involve predicting the numerical labels of an entity at a given seed time. We use Mean Absolute Error (MAE) as our metric and document results in Table~\ref{tab: reg}. Across all datasets and tasks, Rel-LLM consistently achieves the lowest MAE (12.306), highlighting its strong generalization capability across diverse regression problems, including user behavior modeling, item ranking, and entity-based forecasting. Notably, ICL + MLP attains a higher MAE of 13.761, underscoring the challenges LLMs face in performing numerical reasoning over relational databases. In contrast, by integrating graph-based entity reasoning, Rel-LLM effectively captures long-range entity dependencies, enhancing its robustness compared to heuristic approaches and traditional machine learning models.

\subsection{Ablation Studies and Discussion}
\paragraph{Zero-shot Performance.} Traditional models such as LightGBM and RDL rely on supervised training with labeled datasets. However, Rel-LLM can generalize using only its pre-existing knowledge, bypassing the need for manual annotation.
Specifically, Rel-Zero achieves reasonable AUROC scores (63.42) without any labeled data across various datasets, comparable to LightGBM (63.66).
For instance, on the \textsc{rel-event} (\textsc{user-repeat}) task, Rel-Zero achieves an AUCROC of 68.12, which is already close to fine-tuned models like RDL (76.89). On \textsc{rel-stack} (\textsc{user-engagement}), Rel-Zero achieves an AUCROC of 69.46, which is quite high compared to LightGBM (63.39) and is competitive with other baselines. This adaptability makes Rel-LLM highly valuable for real-world applications where labeled data is scarce or expensive to obtain. Nevertheless, there are cases where Rel-Zero struggles more, such as \textsc{rel-event} (\textsc{user-ignore}), where it achieves 61.32, which is far behind RDL (81.62), and the fine-tuned Rel-LLM (83.74). Some failure exploration over many-shot in-context learning for Rel-Zero can be found in Appendix~\ref{app:many-shot}. 
\begin{figure*}
    \centering
    \includegraphics[width=0.95\linewidth]{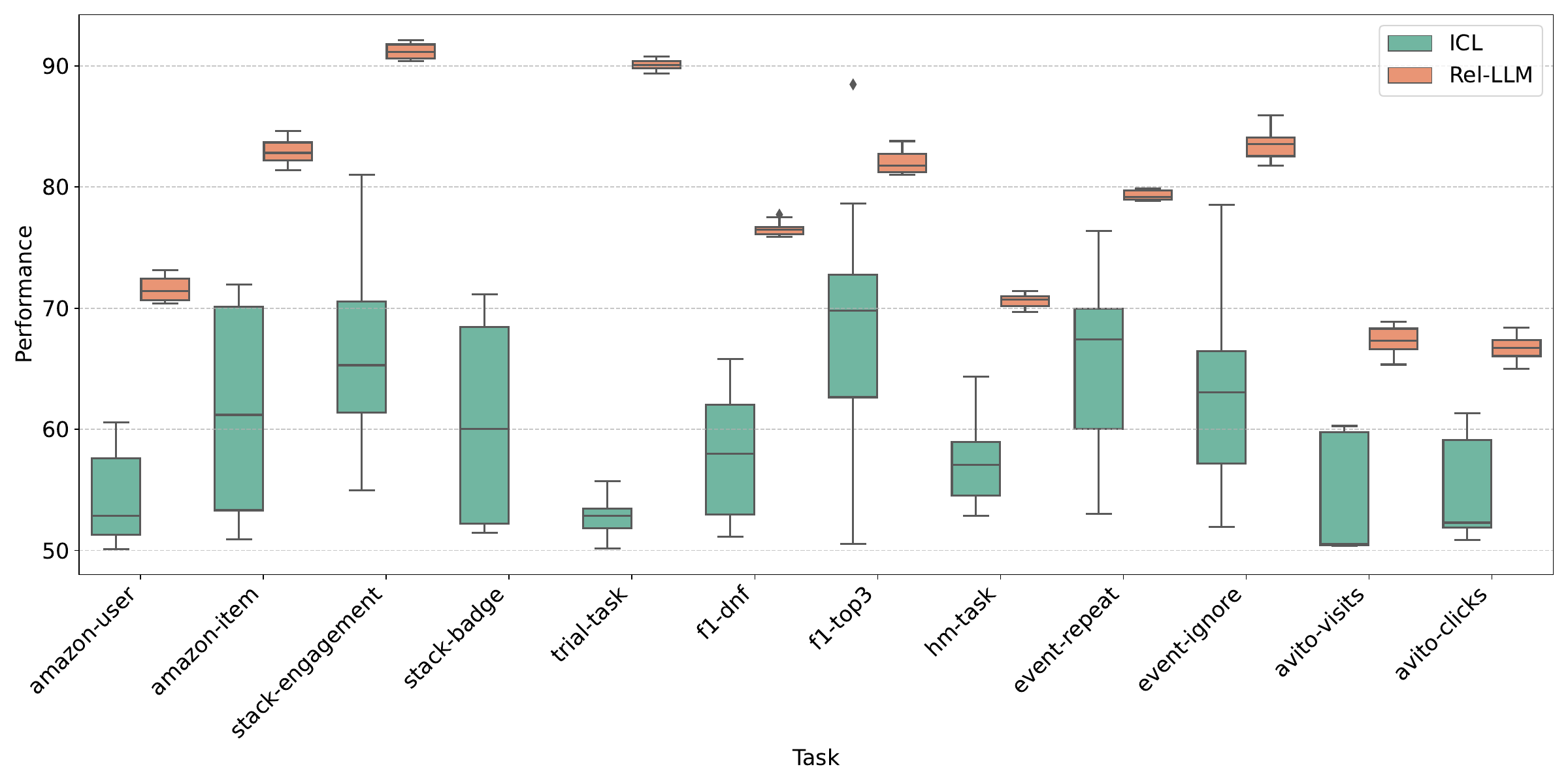}
    \vspace{-1em}
    \caption{Comparision of model prediction variation across different classification tasks.}
    \label{fig:variance}
\end{figure*}
\paragraph{Robustness to In-Context Examples.} While ICL offers a convenient way to forecast an entity’s future behavior without end-to-end training, its predictions are highly variable and sensitive to the order and number of demonstrations. Figure~\ref{fig:variance} illustrates the fluctuations in ICL’s performance across different document parameter settings, such as the number of in-context and related examples. The results indicate that ICL is notably unstable under varying document generation conditions, consistent with prior findings on ICL’s unpredictability~\citep{lu2021fantastically,min2022rethinking,garg2022can}. Furthermore, additional factors, such as the permutation of examples, remain unexplored. In contrast, Rel-LLM demonstrates strong robustness to the randomness of subgraph sampling and variations in in-context examples (see Appendix~\ref{app:in-context}), making it a more reliable and trustworthy choice for real-world applications.

\begin{figure}[t]
    \centering
    \includegraphics[width=1.0\linewidth]{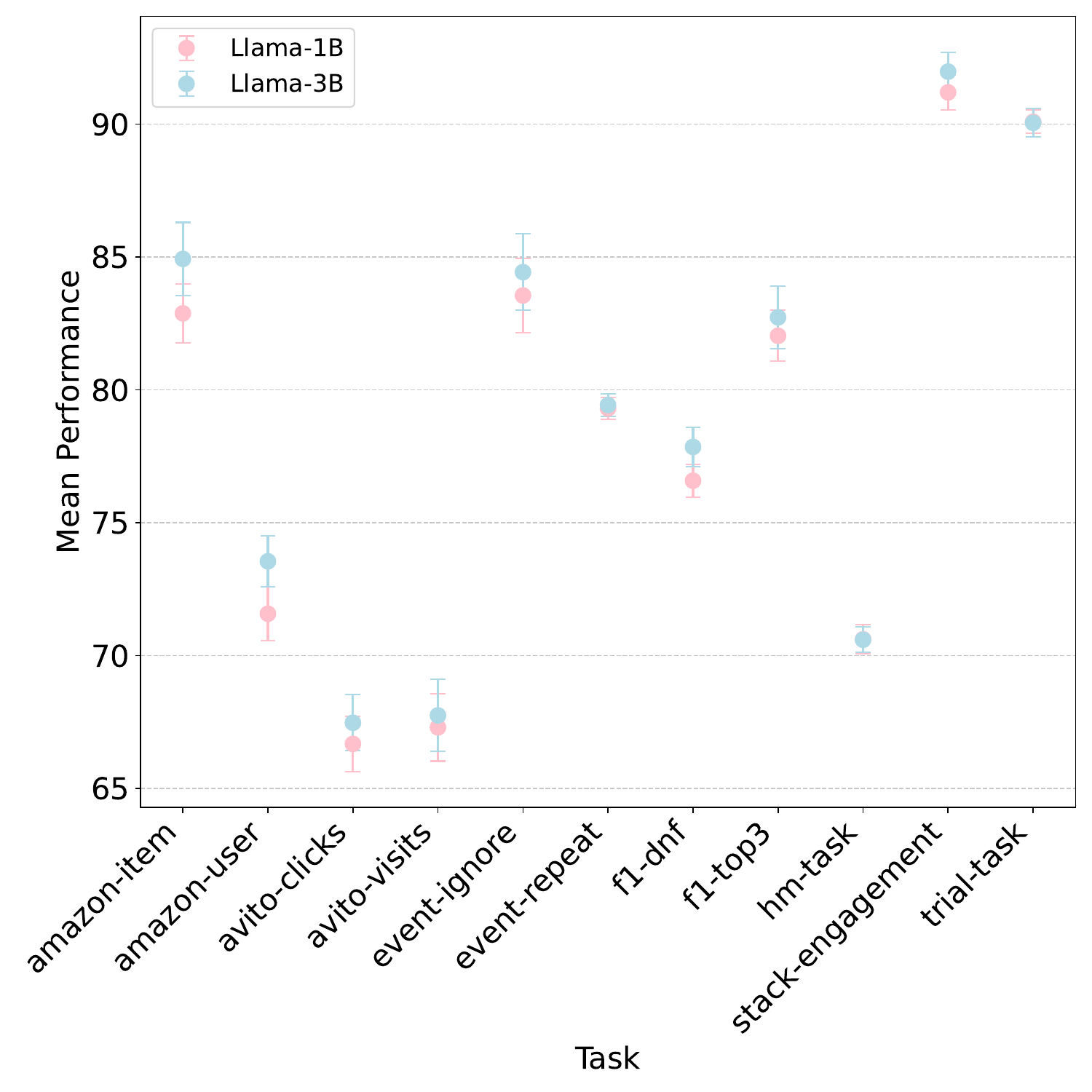}
    \vspace{-1.5em}
    \caption{Ablation on LLMs' model size. }
    \label{fig:model_size}
\end{figure}
\paragraph{Effect of LLM Model Size.} We investigate the impact of scaling up the model size of Rel-LLM to understand whether larger models yield substantial performance improvements. Due to computational resource limitations, we compare the performance of the 1B (billion) parameter version of Rel-LLM with its 3B counterpart, specifically Llama-3.2-3B. Figure~\ref{fig:model_size} illustrates the comparison, revealing that the 3B model generally outperforms the 1B model across most tasks. However, the performance gap between the two models is relatively modest, suggesting that while increasing model size provides benefits, the returns may diminish at this scale. Notably, the 3B model demonstrates a more pronounced improvement in the \textsc{Rel-F1} task, where its larger capacity allows it to better leverage pretraining knowledge, particularly in scenarios requiring complex reasoning or deeper contextual understanding. This indicates that model size plays a critical role in tasks where knowledge utilization and generalization are paramount, though the extent of improvement may vary depending on the specific task requirements.

\section{Related works}
\label{app:related_works}
\paragraph{RDL.} Relational learning seeks to capture inter-entity relationships~\citep{struyf2010relational}, with early methods like inductive logic programming~\citep{de2008logical}, probabilistic logic~\citep{de2008probabilistic}, and relational RL~\citep{zambaldi2018relational} limited by scalability. RDL~\citep{chen2024relational} overcomes this by combining deep learning’s expressiveness with relational reasoning to model non-Euclidean structures, unlike conventional DL optimized for grid-like data~\citep{lecun2015deep}. Earlier approaches~\citep{nickel2011three,perozzi2014deepwalk} often neglected dependencies, prompting the rise of GNNs~\citep{kipf2016semi,chen2021topological,wu2023rethinking,wu2024discovering}. RDL is now key to domains such as social networks~\citep{hamilton2017inductive}, recommender systems~\citep{ying2018graph}, knowledge graphs~\citep{nickel2016holographic}, drug discovery~\citep{rozemberczki2021unified}, fraud detection~\citep{ma2021comprehensive}, and supply chains~\citep{wasi2024graph}.

\paragraph{LLMs for Tabular Data.} Breakthroughs in LLMs facilitate rigorous exploration of tabular data modeling~\citep{fang2024large}, advancing tasks such as prediction~\citep{hegselmann2023tabllm,wang2023unipredict}, tabular data synthesis~\citep{gulati2024tabmt}, question answering~\citep{ye2023large}, and table understanding~\citep{sui2023tap4llm}.
To effectively adapt LLMs for tabular data, researchers have employed techniques focusing on key components like data retrieval~\citep{padhi2021tabular,wang2022transtab}, serialization~\citep{iida2021tabbie,sui2024table}, table manipulations~\citep{liu2023rethinking}, prompt engineering~\citep{chen2022large}, self-supervised learning~\citep{yang2023unitabe,iida2021tabbie}, and end-to-end agentic workflows~\citep{abraham2022tablequery}. 
Despite these strides, challenges remain. Key areas for improvement include reducing bias in tabular data interpretation~\citep{liu2023investigating}, mitigating hallucinations in model outputs~\citep{akhtar2023exploring}, enhancing interpretability~\citep{hegselmann2023tabllm}, and achieving greater computational efficiency~\citep{ruan2024language}.  Furthermore, there has been limited progress in addressing the complexities of relational data within more intricate tabular structures.

\paragraph{RAG on Tables.} RAG~\citep{gao2023retrieval,chen2024benchmarking} enhances LLMs with contextual grounding via data preparation, indexing, retrieval, and augmentation~\citep{gupta2024comprehensive}. In large tables, it reduces processing load using strategies like random sampling (TAPEX~\citep{liu2021tapex}), sliding windows (TabBert~\citep{padhi2021tabular}), row partitioning (TUTA~\citep{wang2021tuta}), truncation (TABBIE~\citep{iida2021tabbie},\citep{qin2022external}), and dimension thresholds (TAGOP\citep{zhu2021tat}). More recent efforts optimize retrieval relevance, with cTBLS~\citep{sundar2023ctbls} using dense dual encoders and DAMO-ConvAI~\citep{ye2023large} leveraging prompt-based row/column selection.


\section{Conclusion} 
We introduced Rel-LLM, a framework combining GNNs and LLMs to enhance relational database processing. By integrating graph prompt tuning, temporal-aware sampling, and self-supervised pretraining, Rel-LLM effectively captures relational semantics while maintaining efficiency. Experiments show noteworthy improvements in accuracy, reasoning, and consistency over baselines, demonstrating its potential for scalable, structure-aware LLM applications. Rel-LLM advances the integration of structured and unstructured data, paving the way for more robust AI systems.

\section{Limitations and Future Works}
\label{app:limitations}
Despite the progress, there is still plenty of room to push the frontier of LLMs for RDL. For instance, it is interesting to investigate a delicate modification of the cross-attention mechanism to better incorporate the graph prompt and LLMs' textual embeddings. 
Additionally, more advanced and larger LLM architectures, such as DeepSeek-R1~\citep{guo2025deepseek}, can be leveraged to improve prompt accuracy due to their stronger pre-existing knowledge and reasoning capacity. Furthermore, exploring how Rel-LLM can be applied to other structured data types, such as knowledge graphs, biomedical datasets, and financial systems, can broaden its impact. 
\bibliography{cite}

\appendix
\newpage
\onecolumn
\section{Prompt Construction Comparison}
\begin{figure}[ht]
    \centering
    \includegraphics[width=0.9\linewidth]{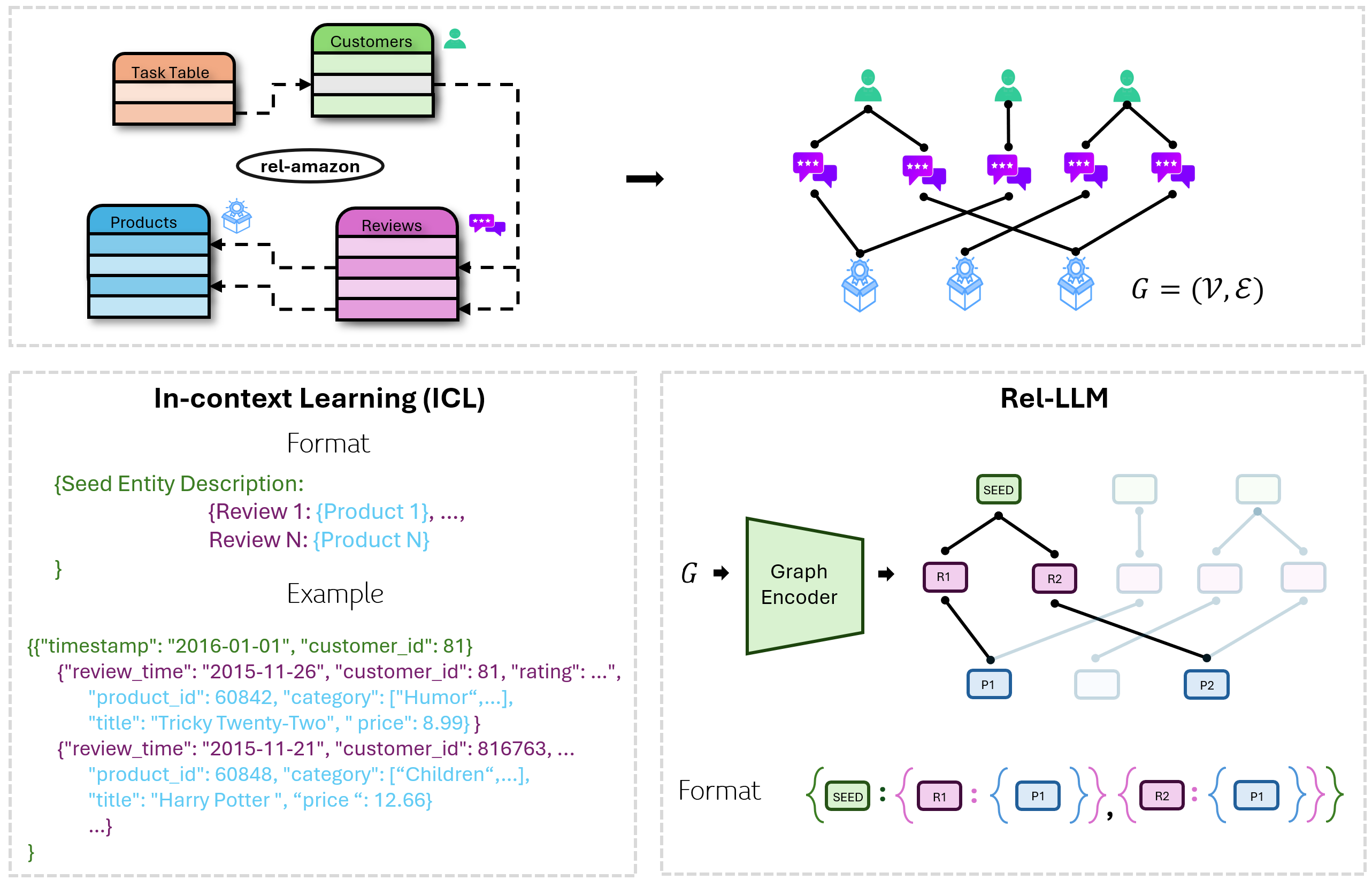}
    \caption{Visualization of different algorithms to compose the task-specific prompts. }
    \label{fig:context}
\end{figure}

The Figure~\ref{fig:context} shows the difference between ordinary ICL and Rel-LLM to formulate the prompts. ICL requires the serialization of tabular data, where great redundancy can be introduced and the document length can be extremely long. As a remedy, Rel-LLM constructs its graph prompt using the graph node embeddings, which are extracted using a graph encoder. 

\section{Experimental Details}
\label{app:exp}
All experiments are conducted on 4 NVIDIA A100 GPUs with 80G memory, mainly relying on Pytorch~\citep{paszke2019pytorch}, PyG~\citep{fey2019fast}, and Huggingface. Pretraining is conducted once. After that, each fine-tuning result is replicated three times, utilizing different random seeds for each run to ensure robustness and reproducibility. The mean results are reported in the main text and the prediction variance is put in the Appendix. 

\subsection{Answer Generation Details}
During pretraining, we leverage plain text generation as the output strategy as the column values of different entities can be of diverse data types. For classification tasks, we exploit the token distribution output strategy, where we set the \emph{maximum\_new\_tokens} to 1mainly focus on the token of "\textsc{Yes}" and "\textsc{No}". We rely on the MLP transformation output strategy for the regression tasks to ensure that the predictions always have float numbers. 

Notably, we do not include zero-shot performance results for Rel-LLM on regression tasks. This is because, after pretraining, LLMs are not inherently designed to generate numerical outputs, such as floating-point numbers, in a structured manner when using a plain text output strategy. Instead, their responses are primarily optimized for generating natural language text. This limitation becomes more pronounced in smaller LLMs, such as Llama 3.2-1B, which struggle to follow explicit instructions in prompts, even when explicitly asked to produce floating-point numbers. As a result, these models often fail to provide reliable numerical outputs for regression tasks, making their zero-shot performance difficult to assess meaningfully.

\subsection{Task Context Details}
Below we elaborate on how we formulate the dataset description $x_{\text{task}}$ and the question prompt $x_{\text{quest}}$. We have explored different prompt templates but observed little performance difference. 
\small
\begin{longtable}{@{}p{2cm}p{2cm}p{11cm}@{}}
    \toprule
    \textbf{Dataset} & \textbf{Task} & \textbf{Description Prompt} \\
    \midrule
    \endfirsthead
    \toprule
    \textbf{Dataset} & \textbf{Task} & \textbf{Description Prompt} \\
    \midrule
    \endhead
    \bottomrule
    \endfoot
    rel-event & user-attendance & This task is to predict how many events each user will respond yes or maybe in the next seven days. \\
    rel-event & user-ignore & This task is to predict whether a user will ignore more than 2 event invitations in the next 7 days. \\
    rel-event & user-repeat & This task is to predict whether a user will attend an event (by responding yes or maybe) in the next 7 days if they have already attended an event in the last 14 days. \\
    rel-amazon & user-churn & This task is to predict if the customer will review any product in the next 3 months or not. \\
    rel-amazon & item-churn & This task is to predict if the product will receive any reviews in the next 3 months or not. \\
    rel-amazon & user-ltv & This task is to predict the \$ value of the total number of products each user will buy and review in the next 3 months. \\
    rel-amazon & item-ltv & This task is to predict the \$ value of the total number of purchases and reviews each product will receive in the next 3 months. \\
    rel-stack & post-votes & This task is to predict how many votes this user's post will receive in the next 3 months. \\
    rel-stack & user-engagement & This task is to predict if a user will make any votes, posts, or comments in the next 3 months or not. \\
    rel-stack & user-badge & This task is to predict if a user will receive a new badge in the next 3 months or not. \\
    rel-avito & user-visits & This task is to predict whether this customer will visit more than one Ad in the next 4 days or not. \\
    rel-avito & user-clicks & This task is to predict whether this customer will click on more than one Ads in the next 4 days or not. \\
    rel-avito & ad-ctr & Assuming the Ad will be clicked in the next 4 days, this task is to predict the Click-Through-Rate (CTR) for each Ad. \\
    rel-f1 & driver-position & This task is to predict the average finishing position of each driver across all races in the next 2 months. \\
    rel-f1 & driver-dnf & This task is to predict if this driver will finish a race in the next 1 month or not. \\
    rel-f1 & driver-top3 & This task is to predict if this driver will qualify in the top-3 for a race in the next 1 month or not. \\
    rel-trial & study-outcome & This task is to predict if the trial in the next 1 year will achieve its primary outcome or not. \\
    rel-trial & study-adverse & This task is to predict the number of affected patients with severe adverse events/deaths for the trial in the next 1 year. \\
    rel-trial & site-success & This task is to predict the success rate of a trial site in the next 1 year. \\
    rel-hm & item-sales & This task is to predict the total sales for an article in the next week. \\
    rel-hm & user-churn & This task is to predict the churn for a customer (no transactions) in the next week. \\
\end{longtable}
\begin{longtable}{@{}p{2cm}p{2cm}p{11cm}@{}}
    \toprule
    \textbf{Dataset} & \textbf{Task} & \textbf{Question Prompt} \\
    \midrule
    \endfirsthead
    \toprule
    \textbf{Dataset} & \textbf{Task} & \textbf{Question Prompt} \\
    \midrule
    \endhead
    \bottomrule
    \endfoot
    rel-event & user-attendance & What is the attendance of user? Give an integer as an answer. \\
    rel-event & user-ignore & Given recent activity and event history, will this user ignore more than 2 event invitations in the next 7 days? Give Yes or No as an answer. \\
    rel-event & user-repeat & Given recent activity and event history, will this user attend an event in the next 7 days? Give Yes or No as an answer. \\
    rel-amazon & user-churn & Based on the customer data provided, will this customer review any product in the next 3 months? Give Yes or No as an answer. \\
    rel-amazon & item-churn & Based on the product data provided, will the product receive any reviews in the next 3 months? Give Yes or No as an answer. \\
    rel-amazon & user-ltv & What is the total dollar value of products this user will buy and review in the next 3 months? Provide a float numerical answer. \\
    rel-amazon & item-ltv & What is the total dollar value of purchases this product will receive in the next 3 months? Provide a float numerical answer. \\
    rel-stack & post-votes & Based on records of activity, how many votes will this user's post receive in the next 3 months? Give an integer as an answer. \\
    rel-stack & user-engagement & Based on records of activity, will this user make any votes, posts, or comments in the next 3 months? Give Yes or No as an answer. \\
    rel-stack & user-badge & Based on records of activity, will this user receive a new badge in the next 3 months? Give Yes or No as an answer. \\
    rel-avito & user-visits & Will this customer visit more than one Ad in the next 4 days? Give Yes or No as an answer. \\
    rel-avito & user-clicks & Will this customer click on more than one Ads in the next 4 days? Give Yes or No as an answer. \\
    rel-avito & ad-ctr & What is the Click-Through-Rate (CTR) for this Ad? \\
    rel-f1 & driver-position & What is the average finishing position of this driver across all races in the next 2 months? Provide a float numerical answer. \\
    rel-f1 & driver-dnf & Will this driver finish a race in the next 1 month? Give Yes or No as an answer. \\
    rel-f1 & driver-top3 & Will this driver qualify in the top-3 for a race in the next 1 month? Give Yes or No as an answer. \\
\end{longtable}
\normalsize 

\subsection{Hyperparameyter Space}
The hidden dimension of Llama $d_l$ is 2048. If a weight decay is applied, then we use the AdamW~\citep{loshchilov2017decoupled} as the optimizer. Otherwise, we leverage Adam~\citep{kingma2014adam} as the optimizer. A Plateau learning rate scheduler is involved with a patience of 100 and a factor of 0.8. 
As for the other important hyperparameters, we use a random search mechanism to find the optimal combination. The entire hyperparameter search space is depicted in Table~\ref{tab:hyper}. The best hyperparameter is selected based on the supervised learning performance on the validation set without pretraining. Then we implement the pretraining and fine-tuning with this fixed set of hyperparameters.
\begin{table}[ht]
    \caption{Hyperparameters setup for Rel-LLM.}
    \centering
    \resizebox{0.72\linewidth}{!}{%
    \begin{tabular}{lll}\toprule
    Hyperparameters Search Space & Symbol & Value \\ \midrule
    \textbf{Training Setup} \\ 
    Epochs & -- & [10, 20, 50, 100]\\ 
    Batch size & -- & [16, 32, 128, 256, 512]\\ 
    Validation steps & -- & [20, 200, 1000, 2000]\\ 
    Learning rate & -- & [1e-4, 5e-5, 1e-6]\\ 
    Weight decay & -- & [0, 1.5e-4]\\ 
    Temporal Strategy & -- & [Uniform, Last]\\ 
    Focusing parameter for focal loss & $\gamma$ & 2.0 \\
    Class balance weight for focal coss & $\alpha_t$  & [0.1, 0.2, 0.4, 0.8, 0.9]\\ 
    \textbf{GNN Architecture} \\ 
    Dropout rate & -- &  [0.1, 0.2, 0.3, 0.4, 0.6, 0.8] \\ 
    Number of GNN layers & $L$ & [2]\\
    Text Embedder & -- & [Glove, MPNet] \\ 
    The output dimension of graph encoder & $d_g$ & [128]\\
    \textbf{Graph Prompt } \\ 
     Number of in-context examples & $n_{\text{inc}}$ & [0, 1]\\
     Number of nested entities & $n_{\text{nest}}$ & [0, 128]\\
     Recursion depth & $\zeta$ & [0, 1] \\
     \textbf{Pretraining Technique} \\ 
    Pretraining epochs & -- & [10, 50, 100]\\ 
    Ratio of masked entities &$ \frac{|\mathcal{V}_{\text{mask}}|}{|\mathcal{V}|}$ & [0.5]\\
    Attribute permutation & $\pi$ & [True, False]\\
     \textbf{LLM Architecture} \\
    Fine-tuning strategy & -- & [Freeze, LoRA]\\ \bottomrule 
    \end{tabular}}
    \label{tab:hyper}
\end{table}

\subsection{GNN Implementation Details} 
Here we provide more details of graph encoder architecture. Message-passing Graph Neural Networks (MP-GNNs)~\citep{gilmer2017neural} are a generic computational framework to define DL architectures on graph-structured data. Given a heterogeneous graph $G=(\mathcal{V}, \mathcal{E}, \phi, \psi)$ with initial node embeddings $\left\{\mathbf{h}_v^{(0)}\right\}_{v \in \mathcal{V}}$, a single message passing iteration computes updated features $\left\{\mathbf{h}_v^{(i+1)}\right\}_{v \in \mathcal{V}}$ from features $\left\{\mathbf{h}_v^{(i)}\right\}_{v \in \mathcal{V}}$ given by the previous iteration. One iteration takes the form:
\begin{equation}
\label{equ:mpnn_iteration}
    \mathbf{h}_v^{(i+1)}=f\left(\mathbf{h}_v^{(i)},\left\{\left\{g\left(\mathbf{h}_w^{(i)}\right) \mid w \in \mathcal{N}(v)\right\}\right\}\right)
\end{equation}
where $f$ and $g$ are arbitrary differentiable functions with optimizable parameters and $\{\{\cdot\}\}$ a permutation invariant set aggregator, such as mean, max, sum, or a combination. Heterogeneous message passing~\citep{schlichtkrull2018modeling,hu2020heterogeneous}is a nested version of Equation~\ref{equ:mpnn_iteration}, adding an aggregation over all incoming edge types to learn distinct message types:
\begin{equation}
\mathbf{h}_v^{(i+1)}=f_{\phi(v)}\left(\mathbf{h}_v^{(i)},\left\{\left\{f_R\left(\left\{\left\{g_R\left(\mathbf{h}_w^{(i)}\right) \mid w \in \mathcal{N}_R(v)\right\}\right\}\right) \mid \forall R=(T, \phi(v)) \in \mathcal{R}\right\}\right\}\right)
\end{equation}
where $\mathcal{N}_R(v)=\{w \in \mathcal{V} \mid(w, v) \in \mathcal{E}$ and $\psi(w, v)=R\}$ denotes the $R$-specific neighborhood of node $v \in \mathcal{V}$. This formulation supports a wide range of different graph neural network operators, which define the specific form of functions $f_{\phi(v)}, f_R, g_R$ and $\{\{\cdot\}\}$.

Given a relational entity graph $G=(\mathcal{V}, \mathcal{E}, \mathcal{T}, \mathcal{R})$ with attached mapping functions $\psi, \phi, \tau$ and initial node embeddings $\left\{\mathbf{h}_v^{(0)}\right\}_{v \in \mathcal{V}}$ and an example specific seed time $t \in \mathbb{R}$, we obtain a set of deep node embeddings $\left\{\mathbf{h}_v^{(L)}\right\}_{v \in \mathcal{V}}$ by $L$ consecutive applications, where we additionally filter $R$-specific neighborhoods based on their timestamp, \emph{i.e.}, replace $\mathcal{N}_R(v)$ with
\begin{equation}
    \mathcal{N}_{\bar{R}}^{\leq t^*}(v)=\{w \in \mathcal{V} \mid(w, v) \in \mathcal{E}, \psi(w, v)=R, \text { and } \tau(w) \leq t^*\},
\end{equation}
which can be realized by the temporal sampling procedure. The formulation naturally respects time by only aggregating messages from nodes that were available before the given seed time $t^*$. The given formulation is agnostic to specific implementations of message passing and supports a wide range of different operators.

\subsection{Class Imbalance Training}
We notice a severe class imbalance in some classification sub-tasks. For instance, in \textsc{driver-dnf} of \textsc{Ref-F1}, the ratios of positives and negatives are 11.96\%/88.04\%, 22.08\%/77.92\%, and 29.49\%/70.51\% for train, validation, and test, respectively. While in \textsc{user-engagement} of \textsc{rel-stack}, the ratios of positives and negatives are 5.0\%/95.0\%, 2.81\%/97.19\%, and 2.74\%/97.26\% for train, validation, and test, respectively. In \textsc{user-clicks} of \textsc{rel-avito}, the ratios of positives and negatives are 3.87\%/96.13\%, 3.52\%/96.48\%, and 1.54\%/98.46\% for train, validation, and test, respectively.
Due to the over-parameterization of LLMs on these tabular data, LLMs usually attain a very high accuracy but a pretty low AUC-ROC. To avoid that, we employ the classic focal loss~\citep{lin2017focal}, which can be written as:
\begin{equation}
    \mathrm{FL}\left(p_\theta\right)=-\alpha_t\left(1-p_\theta\right)^\gamma \log \left(p_\theta\right),
\end{equation}
where $p_\theta$ is the predicted probability of the true (positive) class. $\alpha_t$ is the weighting factor and $\gamma$ is a focusing parameter that adjusts the rate at which easy examples are down-weighted. 

\section{Unsuccessful Attempts}
In the early stage of developing Rel-LLM, we encounter failures and setbacks along the way. We provide our failure experiences here to provide insights and explanations. 

\subsection{Few-shot In-Context Examples}
\label{app:in-context}
To facilitate in-context learning, we include $n_{\text{inc}}$ labeled examples, denoted as $x_{\mathrm{context}}$. Each example consists of an entity $v_i$ from the training set and its corresponding label $y_i$. We enforce a temporal constraint $t_{v_i} < t^*$ for all $v_i$ to maintain chronological validity. Stratified sampling ensures balanced positive and negative examples in binary classification settings where $y_i \in {0,1}$. These examples $\{v_i,y_i\}_{i=1}^{n_{\text{inc}}}$ remain consistent across all documents in a given task to enhance model stability. 

However, we observe that Rel-LLM does not benefit from the inclusion of in-context examples. To elucidate this phenomenon, we refer to the framework proposed by~\citet{pan2023context}, which decomposes the effect of in-context learning into two distinct roles: task recognition (TR) and task learning (TL). Their findings indicate that TR does not improve with increasing model size or additional demonstrations, whereas TL is acquired as model capacity scales. In our case, given that Llama 3.2-1B is relatively small, the provided demonstrations primarily contribute to TR rather than TL. However, following fine-tuning and the explicit inclusion of the task description $x_{\text{task}}$ and question prompt $x_{\text {quest }}$, Rel-LLM has already fully internalized the task. Consequently, the addition of in-context examples has minimal impact. Nonetheless, this conclusion may not hold when employing significantly larger LLMs.

\subsection{Many-shot In-context Learning with Graph Neural Prompt}
\label{app:many-shot}
\begin{figure}[h]
    \centering
    \includegraphics[width=0.9\linewidth]{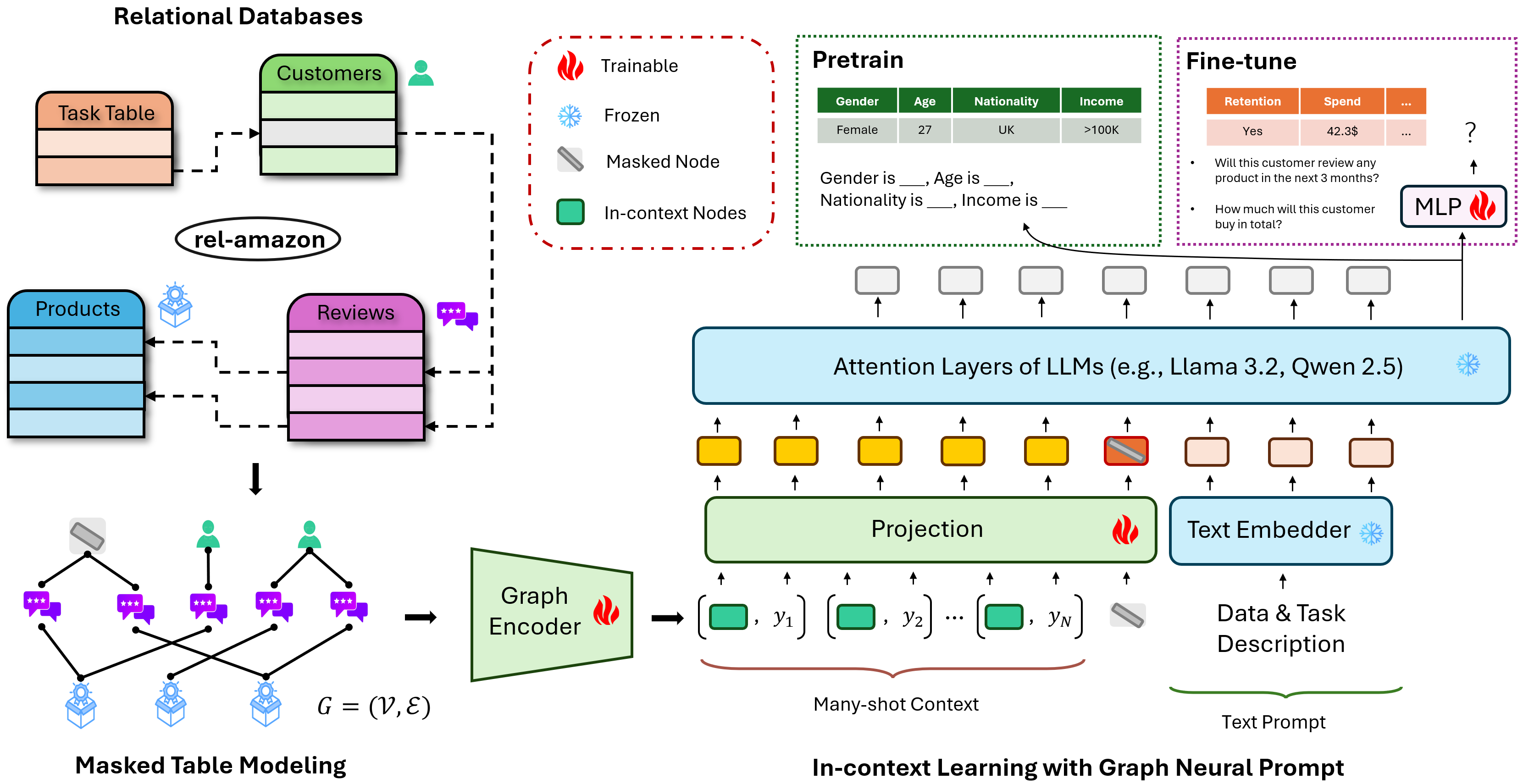}
    \caption{Illustration of many-shot ICL with graph neural prompt.}
    \label{fig:many-shot}
\end{figure}

Recent studies~\citep{agarwal2025many} have shown that LLMs tend to benefit from an increased number of in-context examples, leading to better generalization and task adaptation. However, a major limitation of conventional in-context learning arises when these examples are represented as raw text, as the total context length grows rapidly with the number of demonstrations. This issue becomes particularly problematic given the quadratic computational complexity of self-attention in transformer-based models, which can make many-shot in-context learning infeasible in practice. To mitigate this problem and investigate the effectiveness of many-shot in-context learning in a more scalable manner, we propose replacing textual in-context examples with graph neural prompts (see Figure~\ref{fig:many-shot}). Instead of encoding each example as a sequence of tokens, we represent each demonstration input as a corresponding node embedding. This embedding is then paired with the ground truth label, significantly reducing the number of tokens required to store in-context examples. By leveraging this approach, we can incorporate a much larger number of examples into the model's context window, facilitating extensive in-context learning.

During the pretraining phase, the input context is structured as:
\begin{equation}
    \mathrm{<Seed Node Embedding> + <Dataset Description> + <Task Description>}.
\end{equation}
After pretraining, inference is performed using the following prompt template:
\begin{equation}
\begin{split}
    & \mathrm{ <Seed Node Emebedding> + <Dataset Description> + <Task Desription>}\\
    & \mathrm{+ <Node Embedding of Example 1> + <Label of Example 1> + ...}\\
    & \mathrm{+ <Node Embedding of Example N> + <Label of Example N> + <Seed Node Emebedding>}.
\end{split}
\end{equation}
Furthermore, to ensure a balanced representation across different classes, we consider a many-shot setting where the number of in-context examples per class is approximately equal. Despite these improvements in memory efficiency and scalability, we observe an unexpected decline in the classification performance of Rel-Zero across all tasks when adopting many-shot in-context learning. Specifically, when comparing few-shot settings ($N \leq 16$) to many-shot settings ($N > 16$), we find no significant improvements, contradicting previous findings that suggest LLMs benefit from a higher number of demonstrations.

Upon further investigation, we identify that the model's predictions are highly sensitive to the choice of in-context examples. The performance varies considerably depending on the specific selection of examples, with different random seeds leading to substantial fluctuations. This instability aligns with observations from prior studies~\citep{wydmuch2024tackling}, which report that the order and composition of in-context demonstrations can disproportionately influence LLMs. We hypothesize that this sensitivity may stem from the model's inability to effectively aggregate information across many node embeddings, potentially due to limitations in how LLMs process structured representations compared to textual descriptions. Further research is required to disentangle whether this issue arises from suboptimal prompt construction, insufficient adaptation of GNN-derived embeddings, or inherent limitations in LLMs' ability to generalize from structured input representations.

\subsection{Cell-level Masking}
\begin{figure}[h]
    \centering
    \includegraphics[width=0.95\linewidth]{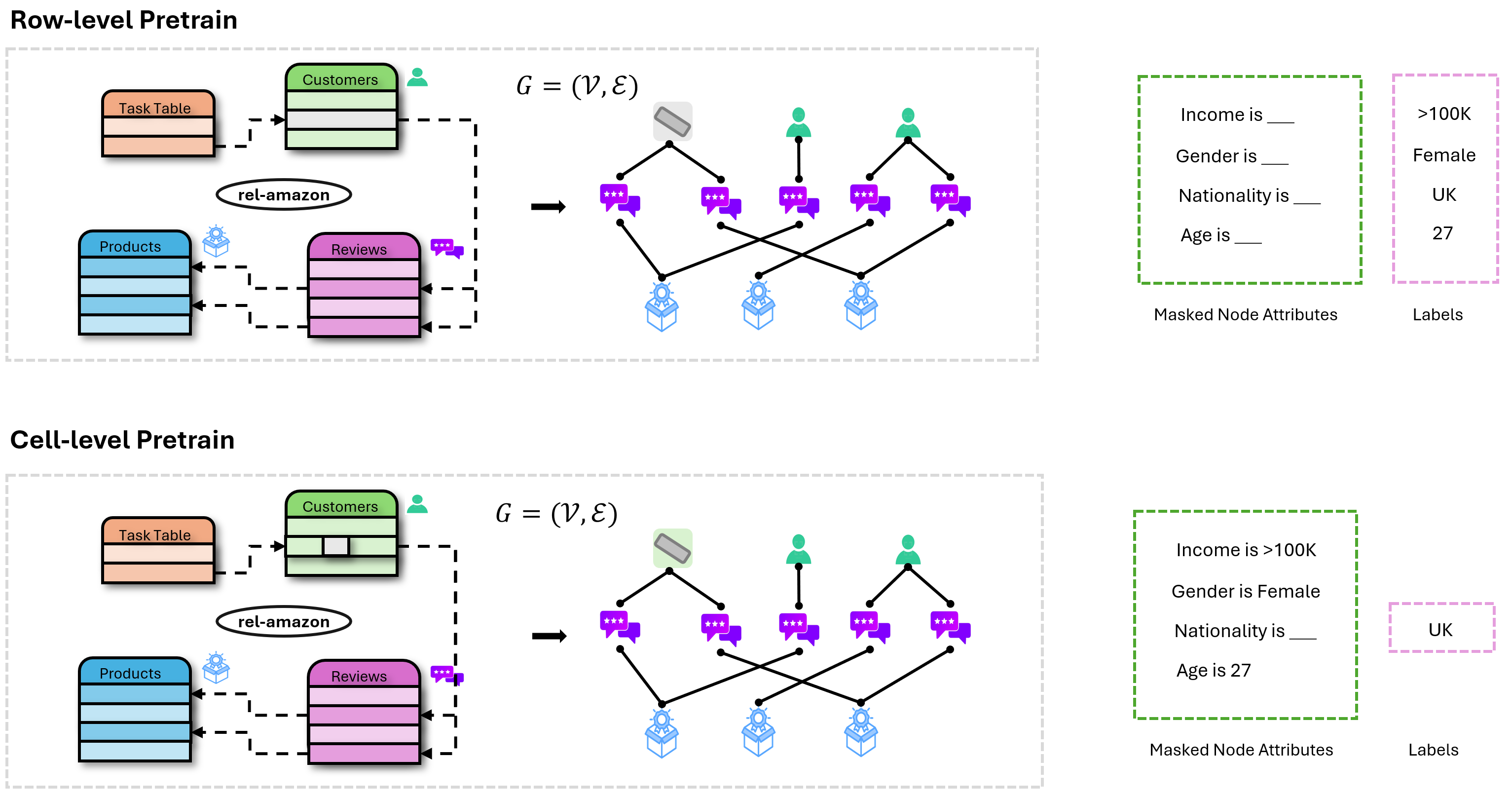}
    \caption{Difference between entity-level and cell-level masking strategies for pretraining. }
    \label{fig:cell-mask}
\end{figure}

Beyond masking entire entities during pretraining, we also explore a more granular approach by selectively masking only certain attributes while keeping others accessible to the large language models (LLMs). This alternative, referred to as cell-level masking, contrasts with the standard row-level masking strategy (see Figure~\ref{fig:cell-mask}). However, empirical results indicate a notable decline in performance under this approach. Specifically, the AUC-ROC of \textsc{Rel-F1} drops from 71.74 to 58.68, while the AUC-ROC of \textsc{Rel-Amazon} decreases from 64.10 to 55.24.

We hypothesize that this performance degradation stems from the inherent dependencies between different attributes within an entity. When only a subset of attributes is masked while others remain visible, the LLMs can often reconstruct the missing values by leveraging correlations across known attributes. This unintended leakage reduces the need for deeper structural reasoning and instead encourages naive inference based on column-wise dependencies. In contrast, entity-level masking disrupts entire relational structures, compelling the model to develop a more holistic understanding of relational graphs rather than relying on shallow pattern matching. Thus, our findings suggest that entity-level masking poses a greater challenge for LLMs, fostering a more robust representation of relational structures.

\section{Additional Results}
Here, we provide the standard deviations over repeated runs for both the entity classification and regression tasks, where the variance is slight and acceptable. 
\begin{table}[ht]
    \centering
    \caption{Standard deviation of performance on the entity regression tasks. }
    \resizebox{1.0\linewidth}{!}{%
    \begin{tabular}{c|cc|c|c|c|c|c|cc} \toprule
        \textbf{Dataset} & \multicolumn{2}{c|}{rel-amazon} & rel-avito & rel-event & rel-f1 & rel-hm & rel-stack & \multicolumn{2}{c}{rel-trial}  \\ \
        \textbf{Task} & user-ltv & item-ltv & ad-ctr & user-attendance & driver-position & item-sales & post-votes & study-adverse & site-success \\ \midrule 
        val & 0.009 & 0.083 & 0.000  & 0.009& 0.025  &  0.001 & 0.002 &  0.389 & 0.018  \\
        test & 0.014  & 0.244& 0.001& 0.008 & 0.132 & 0.002 & 0.002 &0.274 & 0.026 \\ \bottomrule 
    \end{tabular}}
    \label{tab:std_reg}
\end{table}
\begin{table}[ht]
    \centering
    \caption{Standard deviation of performance on the entity classification tasks. }
    \resizebox{0.89\linewidth}{!}{%
    \begin{tabular}{c|cc|cc|cc|cc} \toprule
        \textbf{Dataset} & \multicolumn{2}{c|}{rel-amazon} &  \multicolumn{2}{c|}{rel-avito} &  \multicolumn{2}{c|}{rel-event} &  \multicolumn{2}{c}{rel-f1}  \\ \
        \textbf{Task} & user-churn & item-churn & user-visits & user-clicks & user-repeat & user-ignore & driver-position & item-sales  \\ \midrule 
        val & 0.012 & 0.08 & 0.09  & 0.47 & 3.08 & 0.42 & 2.09 & 4.77  \\
        test & 0.012  & 0.10 & 0.15 & 3.51 & 2.76 & 1.86 & 0.94 & 2.85  \\ \bottomrule         
    \end{tabular}}
    \label{tab:std_cls}
\end{table}
\FloatBarrier
\begin{table}[ht]
    \centering
    \resizebox{0.65\linewidth}{!}{%
    \begin{tabular}{c|c|cc|c} \toprule
        \textbf{Dataset} & rel-hm &  \multicolumn{2}{c|}{rel-stack} & rel-trial  \\ \
        \textbf{Task} & user-churn & user-engagement  & user-bagde & study-outcome \\ \midrule 
        val & 0.017 & 0.16 & 0.14  & 0.96 \\
        test & 0.23  & 0.15 &  0.14 & 2.47  \\ \bottomrule 
    \end{tabular}}
\end{table}

\end{document}